\documentclass{article}
\usepackage[utf8]{inputenc}
\usepackage[normalem]{ulem}
\usepackage{graphicx} % Include figure files
\usepackage{xcolor}
\usepackage{tablefootnote}

\usepackage[font=small,labelfont=bf]{caption}
\usepackage[square,numbers]{natbib}
\usepackage{hyperref}

\title{Strategizing University Rank Improvement using Interpretable Machine Learning and Data Visualization}
\author{Nishi Doshi, Samhitha Gundam, Bhaskar Chaudhury \thanks{Corresponding Author : bhaskar\_chaudhury@daiict.ac.in}\\
\emph{Group in Computational Science and HPC} \\ \emph{DA-IICT, Gandhinagar - 382007 , India.}}
\date{ }
\providecommand{\keywords}[1]{\textbf{\textit{Keywords-}} #1}
\begin{document}
\maketitle

%\author[1]{\fnm{Nishi} \sur{Doshi}}
%\email{201601408@daiict.ac.in}

%\author[1]{\fnm{Samhitha} \sur{Gundam}}
%\email{201601065@daiict.ac.in}

%\author*[1]{\fnm{Bhaskar} \sur{Chaudhury}}\email{bhaskar\_chaudhury@daiict.ac.in}

%\affil[1]{\orgdiv{Group in Computational Science and HPC}\\ \orgname{DA-IICT}, \orgaddress{\city{Gandhinagar}, %\state{Gujarat}, 
%\country{India}}}


\begin{abstract}
Annual ranking of higher educational institutions (HEIs) is a global phenomenon and have significant impact on higher education landscape. Most of the HEIs pay close attention to ranking results and look forward to improving their ranks. However, maintaining a good rank and ascending in the rankings is a difficult task because it requires considerable resources, efforts and performance improvement plan.  In this work, firstly, we show how exploratory data analysis (EDA) using correlation heatmaps and box plots can aid in understanding the broad trends in the ranking data.
Subsequently, we present a novel idea of classifying the rankings data using Decision Tree (DT) based algorithms and retrieve decision paths for rank improvement using data visualization techniques. Using Laplace correction to the probability estimate, we quantify the amount of certainty attached with different decision paths obtained from interpretable DT models. The proposed methodology can aid Universities and HEIs to quantitatively assess the scope of improvement, adumbrate a fine-grained long-term action plan and prepare a suitable road-map.
\end{abstract}

\keywords{University Rankings, Exploratory Data Analysis, Decision Trees, Decision Analysis, Data visualization, Interpretable Machine Learning}
%\end{keywords}



%\maketitle

\section{Introduction}
Despite continuing debate on the pros and cons of university rankings, it can not be denied that university rankings and their influence are a global phenomenon now \cite{gadd, marginson, niilo, goglio, kaycheng, johnes, Myers}. Since rankings provide a mechanism for assessment and accountability of higher educational institutes, it has become a common practise to compare academic and research institutions using different international and national ranking frameworks \cite{grewal,ranksimilarity, wilbers}. Every year, universities are ranked based on their performances which eventually creates a perception that highly ranked institutions are generally more productive in terms of quality research and teaching, and contribute best to society~\cite{shin2011}.
%Ranks of universities have become a major criterion for choosing the university for education. 
The first perception of any educational / research institute is guided by the rank it secures in a national or international ranking framework \cite{ten}. A good rank not only helps the educational institutes in attracting students during admissions but also aids in attracting better research funding, alumni funds, quality faculty, industry projects and better employability of its graduates~\cite{seven, thirtyfour, seventy, seventy-one}. Improvement in rankings are used as an evidence of progress in the academic and research environments when requesting research and development funding from government sources. With the increased worth of higher education, its massification and internationalization, universities are driven to become competitive marketplaces and there is a 
continuous race for excellence~\cite{hazelkorn, grewal, nineteen}. An European University Association (EUA) study on rankings found that universities pay close attention to ranking results even when they are critical of the shortcomings of rankings~\cite{hazelkorn}.
A survey conducted to study the impact of university rankings on universities concluded that most universities have a strong desire to be ranked amongst the top universities, thus suggesting that universities look forward to improving their ranks~\cite{twentyeight}.

Maintaining a good rank and ascending in the rankings is not easy because it requires considerable resources, efforts and planning. To prepare a roadmap for rank improvement, an institute needs to understand the specific ranking model, the important trends in the ranking data, perform a SWOT analysis and quantitatively assess the scope of improvement with available resources. Although several issues have been raised regarding the different ranking models and the indicators used to capture the quality of universities, but still there is no consensus on the best possible framework~\cite{raan}. Besides, data collection and its reliability still remains a major weakness. University rankings are generally determined based on scores that are a congregation of parameters attached to measurable variables. Generally, each ranking framework uses its own set of parameters and the data for individual metrics are condensed into a single final score for determining the rank thereby making it a complex multivariate problem.
Every university ranking framework uses definite rules to rank the universities based on the data collected about their performance in chosen set of parameters. However, recent findings show that there are reasonable similarities between different ranking frameworks, even though each applies a different methodology~\cite{ranksimilarity}.
Research is a parameter that appears across most ranking systems and is considered one of the most important factor~\cite{plosone18}. However, as ranking systems rely on varied quantitative values, it becomes a major challenge for a university to know which parameter greatly affected its ranking. 
An in-depth quantitative analysis and visualization of ranking data (absolute and relative) can help the universities to objectively identify the areas of improvement and layout a framework for improvement. Improvement in rankings might need concentrated attention on underlying issues pertaining to an institute. For instance, a decline in rankings for a parameter related to outreach and inclusivity (gender diversity amongst student, region diversity, economically and socially challenged students) may suggest that the university needs to increase its efforts to reach out to such students, and provide a more welcoming atmosphere.

Various researches related to different issues associated with University Rankings have been performed. Most efforts in the past have focused on assessing the methodology of different ranking systems or examining the impact of such rankings on higher education. Very limited research has been performed to provide a way for a university to improve its rank in a particular ranking framework. Principal Component Analysis and Factor Analysis have been used to understand the importance of the parameters used in the ranking model and concluded that reputation and research output of the university are primary factors that determine the university ranks~\cite{sixteen}. Encouraging academicians for research publication as a strategy to improve international academic ranking has been also examined in the past~\cite{fatt}. A recent study shows how specific changes in strategic direction not only improve a university’s market position but can also contribute to a significant rise in their rank~\cite{dowsett}. A work focused on ranking smaller academic entities such as university departments hints at how difficult it is for an institution to improve its position~\cite{sziklai}.
In this paper, we present a novel idea of classifying the rankings data using Decision Trees, plotting them visually, and propose a methodology to quantitatively assess the scope of improvement and make knowledgeable decisions for rank improvements. 
%The decision tree algorithm is a classifier algorithm that creates a tree-like structure where each node represents a decision rule that is used to partition the data. 
Decision trees include realism in decision analysis by considering the possibility of other chance events and have been successfully used for predictive analysis in several domains ~\cite{decisiontree,thirtyseven, fifty-eight}. To the best of our knowledge, interpretable machine learning (ML) has not been used in analyzing university ranking data or towards higher educational institute rank predictions. The advantage of using decision trees is that it is one of the easier explainable machine learning models and also aids in uncertainty quantification.
Furthermore, we show how visual representation of data obtained from Decision Tress can help universities in easily identifying the scope of improvements in different domains. \\
The rankings has possibly established the belief that improvement is only possible in relation to the performance of other institutes~\cite{brankovic}. Therefore, to quantify the scope of any improvement in this context requires the data associated with a group of institutes which has been ranked by a particular ranking framework.
The National Institutional Ranking
Framework (NIRF) ranking data has been considered in this work~\cite{thirtythree}, however the proposed methodology can be extended to other university ranking system data. 
NIRF ranking framework provides a plethora of scientometric data related to different aspects of educational institutes~\cite{gangan}. The data consists of size-dependent and size-independent parameters, and also several components in terms of quantity and quality  in a meaningful way. Finally, the NIRF model reduces the vast higher education institutional data into a single score. More details related to NIRF is provided in the next section.

The main goal of this work is to demonstrate how universities may use the ranking data to identify opportunities for performance improvement in comparison to other institutions and chalk out a fine-grained long-term action plan for different domains such as academics, research, graduate employability, peer reputation etc. We use data science and data visualization techniques in order to examine the main characteristics of the ranking data, and propose a ML based decision-making strategy for rank improvement. 

%The paper is structured as follows. In Section $2$, we describe the dataset used in this work and important findings from exploratory data analysis (EDA) techniques applied to this data. In Section $3$, we present how the decision tree (DT) algorithm can be applied on the ranking data and discuss the important results. In Section $4$, we provide the details related to DT visualisation based decision making and a detailed analysis of the DT outcomes followed by conclusion in section 5.  

\section{Data : Introduction to the NIRF Rankings}
\label{sec8}
 Every ranking system uses a certain set of parameters to calculate a score and the universities are ranked based on the score they achieve in that Ranking System. THE World University ranking uses indicators that include teaching, research, citations, international outlook and industry income. QS Ranking uses teaching, research impact, reputational standing, student employability, and internationalization factors to rank universities. NIRF uses five parameters to rank universities that include teaching and learning, research and professional practice, outreach and inclusivity, graduation outcome and perception~\cite{thirtythree}. 
It can be observed that all the ranking systems consider similar kinds of factors, hence, without any loss of generality, in this paper, we take into consideration the data of the top $100$ Engineering Institutes in the NIRF Ranking System for the year $2018$ and $2019$. 
% A research study shows that over the years, changes and deviations are observed in the parameters considered by NIRF\cite{twentynine}\textcolor{red}{- why to mention this?}. 
\begin{figure*}[h]
    \centering
    \includegraphics[scale=0.45]{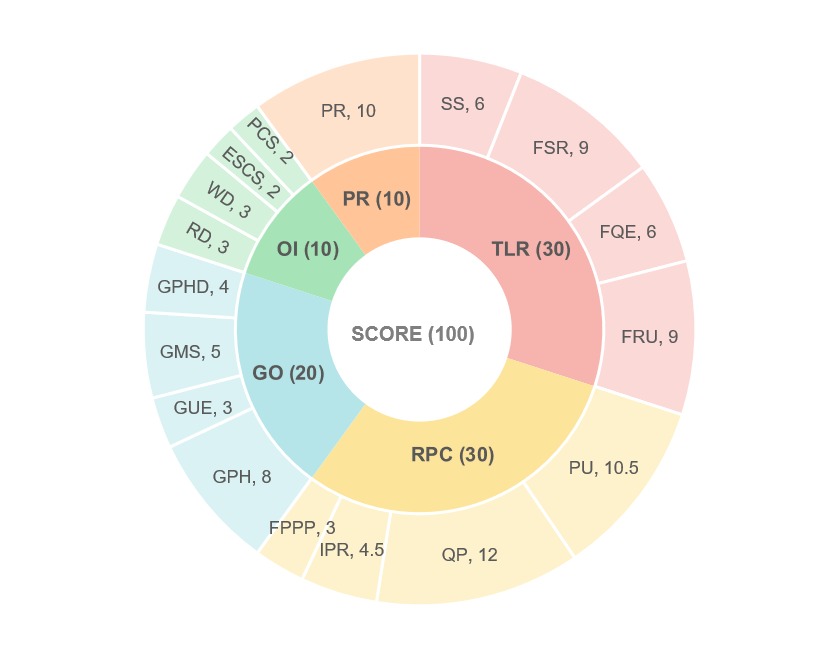}
    \caption{NIRF Ranking Metrics: Pie Chart depicting contribution of Level 1 (Primary heads) and Level 2 (sub-heads shown in outermost circle) parameters towards final NIRF score~\cite{thirtythree}.
    \\
    %Parameter \textbf{TLR} (Teaching, Learning and Resources) includes sub-parameters FRU-final resources \& utility, FQE-faculty metric, SS-student strength, FSR-Faculty Student Ratio; Parameter OI :Outreach and Inclusivity) includes PCS-physically challenged students, ESCS-economically \& socially challenged, WD-woman diversity, RD-outstate/out country students metric, GO-Graduation outcome, GPH-placement \& higher studies, GUE-university exam metric, GMS-median salary metric, GPHD-metric for PhD students graduated, RPC-Research \& Professional Practice, PU-publications metric, QP- the quality of publications, IPR- metric for patents, FPPP-footprint of projects \& professional practice, PP-perception, PR- peer percept0ion} \textcolor{blue}{all weightages (including subparameters) in fraction needs to be provided
    }
    \label{fig4}
\end{figure*}
The NIRF, launched in 2015 by
the Ministry of Human Resource Development
(MHRD), Government of India,
ranks higher educational institutions in the country and are released annually since 2016.
NIRF framework considers 5 parameters to capture the vast complexity of higher
education data and converts it into a single final SCORE (can take a maximum
value of 100) for ranking the institutions. Each of these five parameters, ‘Teaching, learning and resources (TLR)’ ‘Research and professional
practices (RPC),’ ‘Graduation outcomes (GO),’ ‘Outreach and inclusivity (OI)’ and ‘Perception (PR)’, are assigned different weightage (in fraction) for calculating the final ‘SCORE’. 
These five primary parameters are further elaborated through additional sub-parameters, with weights assigned to each sub-parameters.
A pie-chart representing the five parameters and the associated sub-parameters along with their weights are shown in Fig. \ref{fig4}.
For each sub-parameter, a
score is generated from data using suitably proposed metrics, and the sub-parameter scores
are then added to obtain scores for each
primary parameter.

To make the model transparent, relevant data needed to suitably
measure the performance score under each sub-parameter are primarily obtained in two sets which are easily verifiable. First set comprises of data that the institution can easily provide and the second set (primarily research related data) is obtained from third party sources.
For the first set, registered	institutions are invited to	submit the required data through an	Online Data Capturing System	(DCS)~\cite{thirtythree}. For the second set, related to research	output	of applicant institutions,	data are retrieved	
from Scopus	(Elsevier Science) and	Web	of	Science (Clarivate	Analytics)~\cite{thirtythree}. Most of the data are collected for a	three-year period (previous three years of data using a common time window for data collection).
Teaching, Learning and Resources (TLR) includes Student Strength (SS), Faculty Student Ratio (FSR), Financial Resources and Utilization (FRU) as well as Quality of Faculty with PhD and experience (FQE). 
Research and Professional Practice (RPC) includes quantity (PU) and quality of publications (QP) per faculty, Patents Published and Granted (IPR) as well as the funds received for research and consultancy per
faculty (FPPP).
Graduation Outcome (GO) includes Combined metric for Placement and Higher Studies (GPH), Metric for University Examinations (GUE), Median Salary of Graduates (GMS) and Metric for Number of Ph.D. Students Graduated (GPHD).
Outreach and Inclusivity (OI) includes Region Diversity in terms of Percentage of Students from other States/ Countries (RD), Women Diversity (WD), Economically and Socially Challenged Students (ESCS) and Facilities for Physically Challenged Students (PCS).
Perception (PR) quantifies how employers and other academic peers perceive the university based on surveys  conducted over a large category of Employers, Professionals and Academicians. Online feedback system and the Perception	Capturing System are used for online capturing	of	feedback from public and institutional perception (from	Peers and Employers).	
Each primary parameter is first assigned a score out of 100 and then multiplying the value of each parameter with its assigned weight; the sum is calculated which is referred to as the final SCORE of a particular institute. 
In this way, an immensely multi-dimensional problem is condensed into a single score on the
basis of which institutions, irrespective
of size or resources, are finally rank ordered
based on these scores. Table \ref{scoreCalculation} shows real data from $2018$ NIRF Ranking, the final score (SCORE) is calculated by multiplying each parameter with a weight. On the basis of score received, universities are sorted in descending order and ranks are assigned. More details related to source of data, data collection, data verification and ranking methodology can be found at the official website of NIRF~\cite{thirtythree}.

\begin{table*}[!h]
    \centering
    \begin{tabular}{|c|c|c|c|c|c|c|}
    \hline
    Rank &  TLR & RPC  & GO & OI & \vtop{\hbox{\strut Perce-}\hbox {\strut ption}} & \vtop{\hbox{\strut Score = TLR*0.3 +}\hbox{\strut RPC*0.3 + GO*0.2 + }\hbox{\strut OI*0.1 + Perception*0.1}}   \\
    \hline
    $1$  & $93.83$ & $91.44$ & $84.91$ & $63.88$ & $100.00$ & $88.95$\\
    \hline
    $25$  & $67.25$ & $31.43$ & $72.38$ & $54.79$ & $22.59$ & $51.82$\\
    \hline
    $50$  & $55.17$ & $26.58$ & $60.60$ & $41.85$ & $34.67$ & $44.30$\\
    \hline
    $75$  & $54.94$ & $11.61$ & $60.08$ & $46.18$ & $16.93$ & $38.29$\\
    \hline
    $100$  & $57.24$ & $2.53$ & $57.97$ & $57.19$ & $5.97$ & $35.84$\\
    \hline
    \end{tabular}
    \caption{Score calculation for $2018$ Ranked Institutes for Engineering category~\cite{nirf2018}.}
    \label{scoreCalculation}
\end{table*}

\subsection{Exploratory Data Analysis}
To examine the NIRF data, including the scores secured by universities and capture the important trends in the value of primary parameters, we perform EDA~\cite{forty-six} using correlation heatmap analysis and box plot analysis of NIRF $2018$ and $2019$ top $100$ rankings (Engineering category) data~\cite{nirf2018, nirf2019}.

\subsubsection{Correlation Heatmap Analysis}
 Correlation analysis on scientometric data provides very interesting insights and have been used in several studies such as to find the relationship between two university ranking systems~\cite{forty-five}, to investigate the correlation among top 100 universities etc.~\cite{shehata}.
Data collected for our study is not normally distributed, therefore instead of Pearson correlation, Spearman correlation analysis has been used to determine the monotonic relationships among variables~\cite{forty-two}. 
%The correlation coefficient helps us to understand how one variable will change with respect to another. 
The range of correlation coefficients varies between $-1$ to $1$.
The value $1$ indicates a perfect positive relationship among the variables, that is with an increase in one variable, another variable tends to increase. 
We calculate Spearman $\rho$ rank correlation coefficient between the final Score and other parameters such as TLR, RPC, GO, OI and PR using the Equation \ref{spearmaneq}. It is calculated between two features, for example Score and RPC where Score = $(X_1,X_2,...,X_{100})$ and RPC = $(Y_1,Y_2,...,Y_{100})$. Amongst the values of properties, we calculate rank($X_i$) and rank($Y_i$) by simply arranging the values in ascending order and providing rank $1$ to the highest value and rank $100$ to the lowest value. The corresponding difference in the Ranks of $X_i$ and $Y_i$ is represented as $d_i$ and the number of samples are represent as $n$.

\begin{equation}
\label{spearmaneq}
\rho = 1 - \frac{6\sum_{i=0}^{i=n}d_{i}^{2}}{n(n^2-1)}  
\end{equation}

\begin{figure*}[!ht]
    \centering
    \includegraphics[scale=0.40]{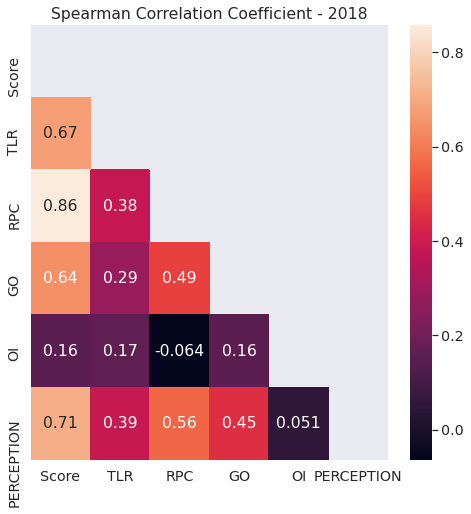}
    \caption{Heat Map of Spearman Correlation for 2018 NIRF Data}
    \label{figure-heatmap-2018}
\end{figure*}

\begin{figure*}[!ht]
    \centering
    \includegraphics[scale=0.40]{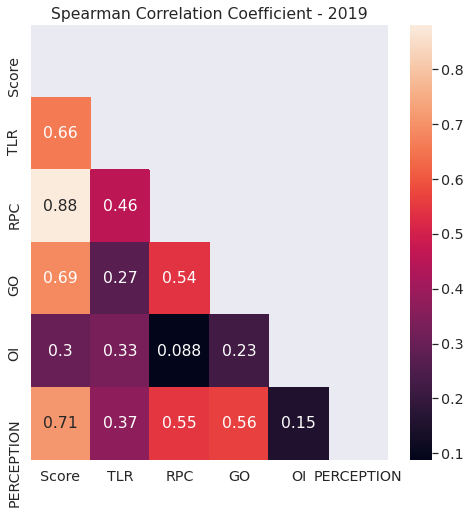}
    \caption{Heat Map of Spearman Correlation for 2019 NIRF Data}
    \label{figure-heatmap-2019}
\end{figure*}

Heat maps representing correlation coefficients for years $2018$ and $2019$ data are presented in Fig. \ref{figure-heatmap-2018} and Fig. \ref{figure-heatmap-2019}. It is observed that the correlation between final score and the parameters considered to calculate this score almost remains same in 2018 and 2019. For example, TLR shows $0.67$ correlation with score in $2018$ and $0.66$ correlation with score in $2019$. Final score shows the highest correlation with RPC around $0.86$ in $2018$ data and $0.88$ in $2019$ data. The final score also shows a $0.71$ correlation with perception. This shows that if a university is ranked amongst the top, it would be a well-known university and would be able to achieve high perception. For both the years, it is observed that RPC shows the highest correlation with final score and OI the least. Additionally, correlation values between different parameters show similar trends in 2018 and 2019. OI and RPC show the least correlation for both years. Thus, we observe that the trend in data values remains the same across years.

Overall a weak correlation is observed amongst the primary parameters (TLR, RPC, GO, OI and PR), signifying that strong monotonic relationship among the variables does not exist. However, there are some interesting trends such as a fair correlation (>.45 in both the years) between RPC, GO and PR exists signifying that institutes which show good performance in research does well in graduation outcomes of students and are also able to create a strong perception. On the other hand, OI has weak correlation with all other sub-parameters. TLR also does not show a strong correlation with other sub-parameters. 
These complex data trends makes it even more difficult for universities to focus on one parameter because if an university decides to improve its score in one parameter, then other parameters may not automatically improve.
For example, an improvement made in TLR has no relation with improvement in OI. 
This makes it very challenging for universities to decide which areas to focus upon or fund to improve their ranking. To acquire more insights about the data, we performed box plot analysis to further understand the scores and trends present in the data of each parameters.

\subsection{Box Plot Analysis}
\label{section-box-plot}
\begin{figure*}[!h]
    \centering
    \includegraphics[scale=1.2]{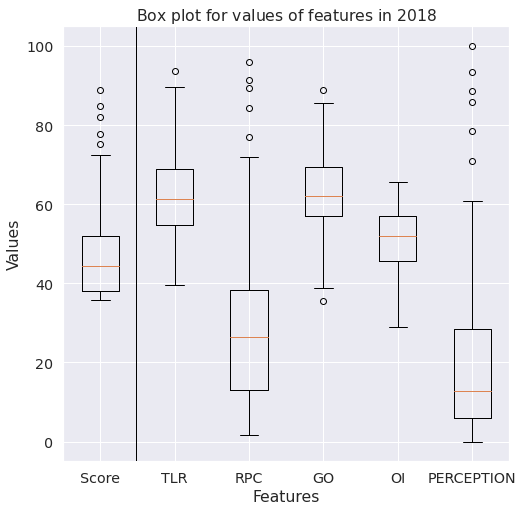}
    \caption{Box plot of parameters considered in NIRF Ranking (2018) 
    \\
    Median Values : Score - $44.285$; TLR - $61.295$; RPC - $26.445$; \\
    GO - $62.22$; OI - $51.995$; Perception - $12.805$}
    \label{fig6}
\end{figure*}

\begin{figure*}[!h]
    \centering
    \includegraphics[scale=1.2]{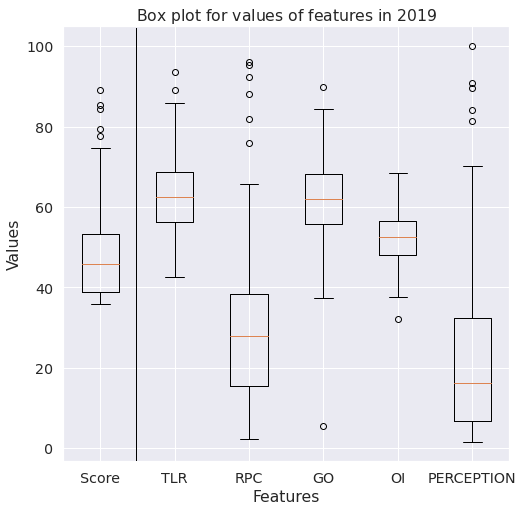}
    \caption{Box plot of parameters considered in NIRF Ranking (2019)
    \\
    Median Value : Score - $45.68$; TLR - $62.435$; RPC - $27.81$; 
    \\
    GO - $61.955$; OI - $52.475$; Perception - $16.13$}
    \label{fig20}
\end{figure*}

Box plot helps to display the data in a standardized format, to determine basic statistical parameters like range, median, outliers in the data and also to detect the important trends in the data~\cite{forty-nine}. The box represents the $50\%$ of the data ($2^{nd}$ and $3^{rd}$ quartile) and the horizontal line in between the box represent the median. The height of the box is defined as the interquartile range (IQR). The horizontal lines above and below the box are known as maximum and minimum of the data defined as $1.5*IQR$ above the $3^{rd}$ quartile and $1.5*IQR$ below the $2^{nd}$ quartile respectively. The dotted circles above or below the maximum and minimum of the data are defined as outliers.
The box plots involving the final score and the 5 parameters (TLR, RPC, OI, GO and PR) are presented in Fig. \ref{fig6} and Fig. \ref{fig20} for years $2018$ and $2019$ respectively. 
%The box plot for both years suggests that the variability in data does not change over the years. 
It is observed that the ranges, median and box sizes of the parameters and the final score is similar in Figs. \ref{fig6} and \ref{fig20}, indicating that the data for $2018$ and $2019$ have almost same statistical characteristics.

From the box plot of the final score, it can be observed that the size of the first quartile (bottom $25\%$ institutes in terms of scores, fourth quartile in terms of ranks) is very small (approximately $3$ points), indicating that the competition is very high in that last $25\%$ of institutes. This means there is a very small difference in the final scores of the institutes lying in the first quartile  compared to other quartiles. Similarly, based on the fourth quartile and outliers data for the final score, it can be seen that the range of scores in the top $25$ universities has a wider distribution.
Based on the score box plot, it can be observed that the top $25\%$ of the universities have scores greater than $50$ points.
The number of outliers with respect to final score, represented using the dotted circles outside the boxes, are the institutes with a very high score, considering that the score is normally distributed. 
In years 2018 and 2019, five institutes are having a very high score considering the distribution of scores of the first $100$ universities.
 
Median as well as the minimum scores for TLR and GO are high compared to other parameters which signifies that there is limited scope of improvement in these two parameters. On the other hand, median score in RPC and PR is below 30 which implies that these two parameters are of concern for most of the institutes and there is a significant scope of improvement.
%We can observe that there are $25$ universities that have RPC greater than 38 points. But there can be no correlation stated from the above, based on which we can conclude that the universities with the highest scores have the highest RPC values.
Box plot of perception (PR) shows that the scores of institutes for this parameter have a large range: $0$ to $60$. However, $75$ percentile of the institutes have a value lesser than $25$, indicating that most universities have a low perception score irrespective of their higher total score.

\begin{table*}[!h]
    \centering
    \begin{tabular}{|c|c|c|c|c|c|c|}
    \hline
    Year & Score & TLR & RPC & GO & OI & \vtop{\hbox{\strut Percep-}\hbox {\strut tion}} \\
    \hline
    $2018$ & $43.5$ & $60$ & $25$ & $60$ & $50$ & $10$\\
    \hline
    $2021$ & $52.5(20.7\% )$ & $90(50\%)$ & $25$ & $60$ & $50$ & $10$ \\
    \hline
    $2021$ & $60(37.9\%)$ & $60$ & $80(220\%)$ & $60$ & $50$ & $10$ \\
    \hline
    $2021$ & $60.1(38.1\%)$ & $77(28.3\%)$ & $50(100\%)$ & $80(33.3\%)$ & $50$ & $10$ \\
    \hline
    \end{tabular}
    \caption{Different possibilities of rank improvement in future (2021) by improving scores of different parameters obtained in current rankings (2018)}
    \label{tab30}
\end{table*}
Although correlation and box plot analysis provides important insights about the overall trend of the ranking data but it does not provide very fine-grained information about how the highly ranked institutes performed in each of these parameters.
There can be multiple ways for a middle or low ranked institute to increase the values of different parameters that determine the final score. 
To understand it better, let us divide the top 100 institutes in four classes or quartiles with each class having 25 institutes based on rank i.e.  class 1 (ranks 1-25), class 2 (ranks 26-50) and so on~\cite{quartile}.
Consider an example, where an institute in $2018$ rankings scored approximately the median score in each of the five parameters as shown in first row of Table \ref{tab30} and falls in class 2 or class 3 (second or third quartile).  Assuming the overall trend in data remains similar to $2018$ (in terms of total score) in future rankings, this institute wishes to improve its rank and ascend to class 1. There are several paths (scores in different parameters) through which it can be achieved, however each path will have different amount of certainty as well as different scope of improvement in different parameters. Let us consider four scenarios.
In the first scenario, shown in $2^{nd}$ row of Table \ref{tab30}, the university puts all its efforts towards improving TLR in next three years (2018-2021) and increases TLR value by $50\%$ in $2021$ and scores for other parameters remain same. This will help to increase the total score by 20.7\%, but the institute will still remain in the class 2 (between rank 25-50). Observing the highest value in TLR ($\approx 90$), it can be stated that the improvement of 50\% may be the maximum for TLR and this may not lead to a very significant increase in final score. Hence, the scope of improvement in TLR is low.

In the second scenario, shown in $3^{rd}$ row of Table \ref{tab30}, the university focuses on RPC and increases RPC value by $220\%$ and scores the same in all other parameters as the previous year. This leads to an overall increase in its score by $37.9\%$ pushing it to class 1. It can be observed from the box plot that RPC also has a high maximum value, hence RPC can be increased to $80$, which would lead to a greater score. Hence, RPC has more scope of improvement compared to TLR. However, such improvements cannot be made on a very short time scale and requires long term strategy.

In the third scenario, shown in $4^{th}$ row of Table \ref{tab30}, the institute puts in efforts to increase TLR, RPC and GO by $28.3\% ,100\% , 33.3 \% $ respectively. It can be observed that, unlike the first and second scenario where all efforts have been put to improve one parameter, the institute can divide the resources into combinations of various parameters to improve the score and move to class 1. As can be observed from the box plot, in this case, the values of TLR, RPC and GO are just a little above the $3^{rd}$ quartile and hence can be achieved with more certainty compared to first two scenarios.

As observed above, increasing score in combination of parameters may help universities to increase its overall score and hence rank in the subsequent years. As score is calculated as a combination of different parameters, it is difficult to visualize this multivariate data and ascertain the ways to improve the ranking in subsequent years. Gestalt principle suggests that the whole is greater than the sum of its parts~\cite{gestalt}, and to interpret the whole we need to visualize and understand the parts very effectively.
We propose to use decision trees to visualize the patterns in ranking data in a single diagram, perform decision analysis, and formulate a plan for improvement.

\section{Method: Decision Trees for Classification of Institutes}

A decision tree (DT) is an interpretable ML model established on binary trees that learns the connection between observations in a training dataset, represented as feature vectors $\textbf{V}$ (in our study TLR, RPC, GO, OI and PR) and target values $\textbf{Z}$ (in this study target classes as shown in Table \ref{table-4class-ranges}), by encapsulating training data into a binary tree consisting of interior and leaf nodes. 
A DT partitions the feature space into classes of observations that share matching target values and each leaf node represents one of these classes (or groups). Any path, starting from the root of the DT reaches to a specific leaf node after passing through a sequence of decision nodes, where the decision node compares the value $v_i$ of a single feature, with a particular split point value learned during training.
The decision node's feature and the split point are selected during training to divide the observations into left and right subgroups to maximize similarity so that most or all targets are of similar class in a subset. The left subset has observations whose $v_i$ feature values are all less than the split point and the rest goes to the right subset. Tree construction continues recursively by generating decision nodes for the left and the right subsets until some stopping criterion is reached. Decision trees are very helpful in constructing different options (or courses of action or paths) and analyse possible outcomes of picking those options. In this study, we are interested in understanding the rules (or decisions) which leads to a specific path (leading to a specific class).
% The visualization of the Decision Tree discussed in the next section helps in decision analysis based on the data fed into the decision tree.
Splitting criteria mentioned above is selected on the basis of DT algorithm used for construction. We use Classification And Regression Trees (CART) and C4.5, algorithms for building a decision tree. CART constructs binary tree using Gini Impurity (GI) as node splitting criteria, splitting a node into two child nodes repeatedly~\cite{fifty-one, breiman2017classification}. ID3 algorithm uses Information Gain (IG) as node splitting criteria~\cite{eleven}. C4.5 algorithm is similar to ID3 algorithm but overcomes few drawbacks of ID3 algorithm such as use of continuous data, pruning the tree after its construction.
%We perform prediction using CART and C4.5 algorithms and visualize decision tree constructed using the algorithm that provides better prediction results. 

Construction of decision tree involves following steps : 
\begin{enumerate}
    \item For every parameter in consideration, split the data in two parts and calculate GI or IG for every split.
    \item Choose the parameter which gives the best impurity reduction after split based on splitting criteria and split the dataset into two subsets.
    \item Repeat Step 1 and 2 on every subset that is generated till leaf node is reached which has data belonging to one class only.
\end{enumerate}
We use decision trees for both prediction and visualization. By using predictive power of decision tree, we show that $2019$ NIRF rankings can be predicted using $2018$ rankings data. Thereafter, we construct a decision tree and visualize it to study $2018$ rankings, and show how universities can create an action plan to improve its rankings. A detailed analysis of visualized decision tree is given in Sec. \ref{decisionTreeVisualization}.

\begin{table*}[!h]
     \centering
    \begin{tabular}{|c|c|}
    \hline
    Class/ Quartile & Range of ranks in this class \\
    \hline
    Class $1$ & Ranks $1$ to $25$\\
    \hline
    Class $2$ & Ranks $26$ to $50$\\
    \hline
    Class $3$ & Ranks $51$ to $75$\\
    \hline
    Class $4$ & Ranks $76$ to $100$\\
    \hline
    \end{tabular}
    \caption{Class Division and Range of Ranks in each class}
    \label{table-4class-ranges}
\end{table*}

% \subsection{ID3 Algorithm}
% ID3 Algorithm was developed by Ross Quinlan\cite{eleven}. Using information gain calculated based on entropy; ID3 algorithm constructs a decision tree. The equation for entropy is given in Eq. \ref{eq4}.
% \begin{equation}
% \label{eq4}
%     E(P) = \sum_{i=1}^{i=n}-p_{i}log_{2}p_{i}
% \end{equation}
% where
% \begin{itemize}
%     \item $E(P)$ = Entropy over all classes
%     \item $p_i$ = Probability of a particular object classified in particular class.
%     \item $n$ = Total number of classes
% \end{itemize}
% Equation for Information Gain is given in Eq. \ref{eq7}.
% \begin{equation}
%     \label{eq7}
%     IG(p,T) = E(p) - \sum_{i=1}^{i=n}p_i E(p_j) 
% \end{equation}
% where 
% \begin{itemize}
%     \item $IG(p,T)$ = Information Gain for test T and position p
%     \item $E(p)$ = Entropy over all classes
%     \item $n$ = Number of classes
%     \item $p_i$ = Probability that $i^{th}$ sample belongs to a certain class. 
% \end{itemize}
% It is the underlying algorithm which further led to the development of CART algorithms. One of the major drawbacks of ID3 algorithm is that it becomes computationally expensive to use the algorithm with continuous variable inputs as it generated many trees\cite{twelve}.

% \subsection{CART Algorithm}
% The decision tree on every node selects one feature at a data point on which the data will be divided. The feature that best splits the data is used as the deciding factor of the split. There are two splitting algorithms used by CART: Gini Index and Information Gain to split the data. \\
\subsection{Gini Index}
\label{sectiongini}
Gini Index determines the purity of class after the splitting along a specific feature. Gini Index for university ranking dataset $URD$ containing observations belonging to each class is calculated by subtracting the sum of probabilities of observations classified per class using Eq. \ref{ginidataeq} 
%presents the mathematical formula to calculate Gini Index for $URD$.
\begin{equation}
\label{ginidataeq}
    Gini(URD) = 1 - \sum_{i=1}^{i=4}p_i^2
\end{equation}

Gini Index, when $URD$ containing $100$ samples is split on feature $RPC$ into two datasets $URD_1$ and $URD_2$ containing $n1$ and $n2$ samples respectively, is calculated using Eq. \ref{ginifeature}.
\begin{equation}
    \label{ginifeature}
    Gini_{RPC}(URD) = \frac{n_{1}}{100}Gini(URD_{1}) - \frac{n_{2}}{100}Gini(URD_{2})
\end{equation}

Thus, by splitting the dataset $URD$ on $RPC$, the reduction in impurity is calculated as the difference between Gini Index of $URD$ calculated in Eq. \ref{ginidataeq} and the Gini Index of datasets obtained after split ($URD_1$ and $URD_2$) calculated in Eq. \ref{ginifeature}. 
%Eq. \ref{ginifinal} shows the mathematical formula that reduces the impurity for the $RPC$ feature. 
\begin{equation}
\label{ginifinal}
    Gini(RPC) = Gini(URD) - Gini_{RPC}(URD)
\end{equation}
Similarly, impurity reduction for TLR, GO and OI are calculated and the feature that results in a maximum increase of purity after split is used to split the data into two sets $URD_1$ and $URD_2$. Subsequently, the algorithm selects the feature which divides $URD_1$ and $URD_2$ separately in a way that purity increases on dividing the data as per the feature.

\subsection{Information Gain (IG)}
\label{informationGain}
Information gain is calculated using entropy. Entropy quantifies impurity or randomness in data~\cite{forty-seven}, and its value ranges between $0$ and $1$. When there is no impurity in data, the value of entropy is $0$. Entropy is calculated by summing the probability of each class times the log probability of the same class. Eq. \ref{entropyeq} is used to calculate entropy over all classes for the $URD$ dataset.
\begin{equation}
\label{entropyeq}
    E(URD) = \sum_{i=1}^{i=4}-p_{i}log_{2}p_{i}
\end{equation}

Information Gain measures the reduction in impurity before and after the split using a specific feature. Information Gain, computed using Eq. \ref{igentropy}, is the difference between entropy of the class and conditional entropy of the class and selected parameter~\cite{forty-eight}. E(URD) is entropy of class and $pi*E(URD_i)$ is the conditional entropy.
\begin{equation}
    \label{igentropy}
    IG(p,URD) = E(URD) - \sum_{i=1}^{i=4}p_i E(URD_i) 
\end{equation}

The decision tree selects the feature which results in maximum information gain that is maximum impurity loss as calculated in Eq. \ref{igentropy}. 

\subsection{Quadratic Weighted Kappa Score}
\label{quadratickappa}
% Eq. \ref{kappaeq} provides mathematical formula to calculate Cohen's kappa score.
% \begin{equation}
% \label{kappaeq}
%     \kappa = \frac{p_{o} - p_{e}}{1 - p_{e}}
% \end{equation}
To evaluate the predictive performance of the Decision Tree, we use the Quadratic Kappa Value as an evaluation metric.
Quadratic weighted kappa score is calculated by using three tables: weights table which measures disagreement between two categories, observed table which captures the predicted classification of test dataset and expected table which shows the actual classification of test dataset~\cite{fifty}. The weights indicate the seriousness of the disagreement. The weight matrix is calculated such that it provides minimal weight to minimal disagreement and maximum weight to maximum possible disagreement. Weights are allocated to each cell of the weighted table by calculating the square of the distance between the actual class and predicted class divided by the $1$ less than the number of classes. Table \ref{tabweights} shows weighted matrix that is constructed using Eq. \ref{weighteq} for $4$ class classification problem. The observed matrix should be constructed based on model results. The expected matrix should be constructed by taking into account the actual classification of the test dataset.
\begin{equation}
    \label{weighteq} w_{ij} = \frac{(i-j)^{2}}{3}
\end{equation}

\begin{table*}[!h]
    \centering
    \begin{tabular}{|c|c|c|c|c|}
    \hline
    Predicted Class / Actual Class & Class $1$ & Class $2$ & Class $3$ & Class $4$ \\
    \hline
    Class $1$ & $0$ & $0.11$ & $0.44$ & $1$\\
    \hline
    Class $2$ & $0.11$ & $0$ & $0.11$ & $0.44$\\
    \hline
    Class $3$ &$0.44$ & $0.11$ & $0$ & $0.11$\\
    \hline
    Class $4$ & $1$ & $0.44$ & $0.11$ & $0$\\
    \hline
    \end{tabular}
    \caption{Weighted matrix used while calculating Quadratic Kappa Score for $4$ class classification}
    \label{tabweights}
\end{table*}
  Quadratic kappa value for $4$ classes is calculated using Eq. \ref{kappaeq} using weighted matrix $W$, the observed matrix $X$ and expected matrix $M$.
\begin{equation}
    \label{kappaeq} 
    \kappa = 1 - \frac{\sum_{i=1}^{i=4}\sum_{j=1}^{j=4}w_{ij}x_{ij}}{\sum_{i=1}^{i=4}\sum_{j=1}^{j=4}w_{ij}m_{ij}}
\end{equation}

% 0.         0.11111111 0.44444444 1.        ]
%  [0.11111111 0.         0.11111111 0.44444444]
%  [0.44444444 0.11111111 0.         0.11111111]
%  [1.         0.44444444 0.11111111 0.        ]]

% [0.     0.0625 0.25   0.5625 1.    ]
%  [0.0625 0.     0.0625 0.25   0.5625]
%  [0.25   0.0625 0.     0.0625 0.25  ]
%  [0.5625 0.25   0.0625 0.     0.0625]
%  [1.     0.5625 0.25   0.0625 0.    ]

The value of quadratic kappa ranges from $-1$ to $1$~\cite{five}. The model used for prediction is considered poor if $\kappa \leq 0.4$, fairly good if $0.4 \leq \kappa \geq 0.75$ and excellent if $\kappa \geq 0.75$~\cite{fifty-three}. Table \ref{tabweights} shows, if an observation of Class $1$ is classified in Class $3$ then it is given more penalty compared to it being classified in Class $2$. As we deal with ranking data, the evaluation of the predictive performance of our model should depict how nearer is the predicted rank to its actual rank which is quantified using the quadratic kappa score. 
%Many predictive classification problems like predicting the severity of spinal pain where data of severe pain classified into the category of mild pain should be penalized more compared to the observation being classified into mild pain category, predicting the scholarly impact of publications using peer review and bibliometrics use Quadratic Kappa value as an evaluation metric~\cite{fifty-five, fifty-six}.

% To reflect the degree of disagreement, kappa can be weighted, so that it attaches greater emphasis to large differences between ratings than the small differences
%%%%%%%%%%%%%%%%%%%%%%%%
%%%%%%%%%%%%%%%%%%%%%%%%
\subsection{Decision Tree Construction}
\label{decision-tree-generation}
\begin{figure*}[!h]
    \centering
    \includegraphics[width=12cm, height=6cm]{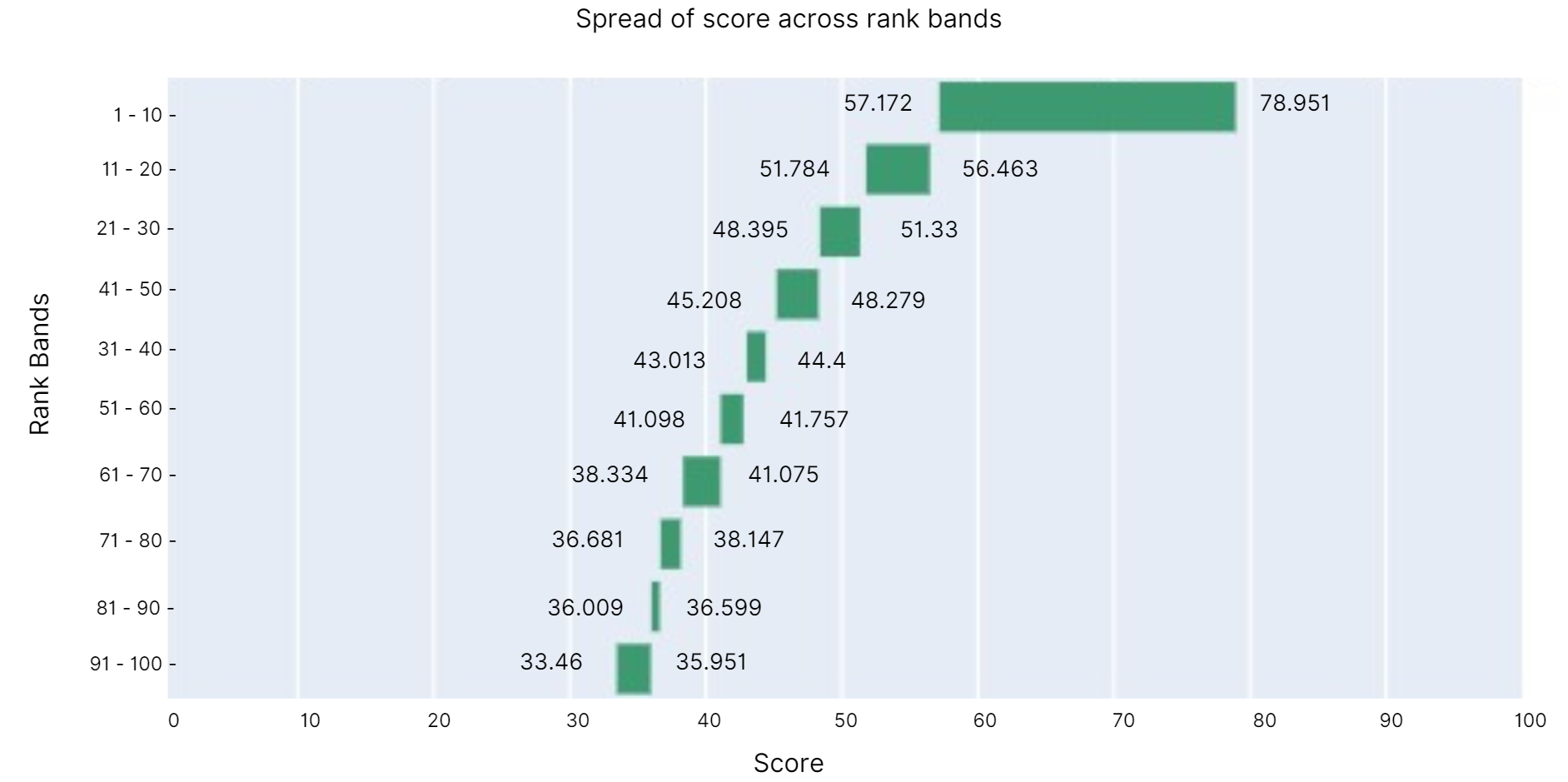}
    \caption{Spread of scores across rank bands for 2018 NIRF Rankings Data (excluding perception (PR) score)}
    \label{fig-scores-acrss-rank-bands}
\end{figure*}

We construct a Decision Tree (DT) Classifier that classifies the samples into different ranking categories (classes).
Before applying the DT Algorithm on the data, we divide universities into $4$ classes based on the ranks as shown in Table \ref{table-4class-ranges}. This division of universities into $4$ classes ensures that each class has a reliable number of samples which can help the decision tree to learn the patterns in the data. Samples can be divided into any number of classes based on the availability of samples. Dividing 100 samples (ranks 1 to 100) into $4$ classes assures every class contains $25$ samples each. In our analysis, we consider TLR, RPC, GO and OI as the features to train the decision tree. We exclude PR (perception) as a feature while constructing the decision tree because its value is based on surveys and opinions of others and it may be quite challenging for an institute to make specific short-term strategies to improve this parameter significantly. Additionally, from the Box Plot analysis discussed in Sec. \ref{section-box-plot}, we observed that the values of parameter perception are lower than $10$ for more than $75\%$ of the institutes even though the rank of institutes are very good. Moreover, we found that Spearman correlation between the rankings of institutes with and without the perception score is $0.989$, indicating that the rankings of institutions without perception are close to the actual rankings. Thus, we consider TLR, RPC, GO and OI as $4$ features in our decision tree construction and our score is determined using only the 4 features.  Fig. \ref{fig-scores-acrss-rank-bands} represents a waterfall chart depicting the spread of scores across rank bands of $10$, calculated without taking perception into account. We found that statistical characteristics of the data in terms of rank bands almost remains same after removing the perception parameter from total score calculation. For subsequent sections, we use the score calculated taking into account TLR, RPC, GO and OI for our experiments and explanations.

The DT is constructed with the training data i.e. $100$ samples from $2018$ NIRF data (the first $100$ ranked institutes in engineering category) and tested using $2019$ ranking data to ensure the reliability of decision tree paths to make decisions for the subsequent year. 
%The DT is trained using the $100$ samples from $2018$ NIRF data (the first $100$ ranked institutes in engineering category) and tested using 2019 data to gain confidence about the predictive analysis of the decision tree.
By constructing a decision tree using $2018$ data, we are visualising the common feature routes taken by various institutes to be part of a class (ranking category) i.e, many institutes of class $1$ (ranked $1$-$25$) may have parameters greater than a certain value in common, hence determining a plausible rule that can be followed by other universities to reach into class $1$. 

\begin{figure*}[!h]
     \centering
     \includegraphics[width=8cm,height=6cm]{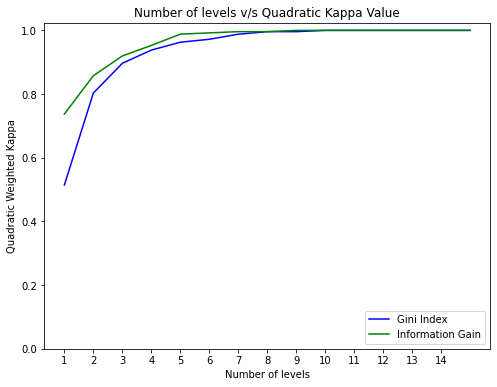}
     \caption{Number of levels vs Quadratic Kappa metric to perform pruning on decision trees}
     \label{fig33}
\end{figure*}

We construct 2 decision trees using different splitting algorithms, one using Gini Index and other using Information Gain to evaluate the performance of both the algorithms.
The final tree obtained from DT algorithm that classifies all of the samples into distinct classes often leads to a lot of levels which makes it difficult to interpret the tree. Hence we define a stopping criteria to limit the number of levels making it more interpretable and easy to visualize. 
%We aim to choose a sub-tree taking into account the complexity(size) of the tree and the kappa metric of the tree. 
We define the stopping criteria, by choosing the number of levels where the accuracy metric of the tree does not change much and the tree is interpretable. In our experiment, we define the stopping criteria as 5 levels because the kappa value almost remains constant after 5 levels as shown in Fig. \ref{fig33}, and on visualizing the tree we find that most of the samples (institutes) are being distinctly classified in the leaf nodes at level 5 itself, as observed in Fig. \ref{fig1}.  

 We tabulate the evaluation metrics obtained from both DT algorithms in Table \ref{table-results-4level}. We observe, the tree constructed using Information Gain as splitting criteria has a greater quadratic kappa score value of $0.902$. The accuracy obtained with the IG splitting algorithm is also high ($75\%$). Table \ref{table-confusion-matrix} shows confusion matrix created by results obtained from IG splitting algorithm. The tables shows that the misclassification happens only by $1$ class difference. The decision tree for $2018$ year does not classify data of Class $1$ into Class $3$ or Class $4$. Similarly, no university present in Class $2$ in $2019$ year is classified as Class $4$. This shows that the decision tree constructed on $2018$ data (present data) is reliable to predict the trends present in $2019$ data (future data). 
%Thus the decision tree constructed with the IG splitting algorithm gives us the confidence that the decision patterns to predict the category of the rank does not change over $2019$. 
Hence, the analysis done on the tree constructed from current data can be used with desirable confidence to make strategies for score improvement in subsequent years.

\begin{table*}[!h]
    \centering
    \begin{tabular}{|c|c|c|}
    \hline
    Splitting Algorithm & Accuracy & Quadratic Kappa\\
    \hline
    Gini Index & $0.66$ & $0.808$ \\
    \hline
    Information Gain & $0.75$ & $0.90234$ \\
    \hline
    \end{tabular}
    \caption{Results for training $2018$ data in decision tree and testing on $2019$ data by dividing data in $4$ classes}
    \label{table-results-4level}
\end{table*}

%\begin{table}[!h]
    %\centering
    %\begin{tabular}{|c|c|c|}
   % \hline
  %  Splitting Algorithm & Accuracy & Quadratic Kappa\\
  %  \hline
 %   Gini Index & $0.83$ & $0.85279$ \\
 %   \hline
%    Information Gain & $0.85$ & $0.86545$ \\
%    \hline
%    \end{tabular}
%    \caption{Results for training $2018$ data in decision tree and testing on $2019$ data by dividing data in $3$ classes}
%    \label{table-results-3level}
%\end{table}

% [[23  2  0  0]
%  [ 5 20  0  0]
%  [ 0 11  8  6]
%  [ 1  2  7 15]]
%  0.66000
% 66/100

%  [[22  3  0  0]
%  [ 3 21  1  0]
%  [ 0  5 12  8]
%  [ 0  0  5 20]]
% 0.75
% 75/100

\begin{table*}[!h]
\centering
\begin{tabular}{|c|c|c|c|c|}
     \hline
     Actual v/s Predicted Class & Class $1$ & Class $2$ & Class $3$ & Class $4$ \\
     \hline
      Class $1$ & $22$ & $3$ & $0$ & 0 \\
      \hline
      Class $2$ & $3$ & $21$ & $1$ & $0$ \\
      \hline
      Class $3$ & $0$ & $5$ & $12$ & $8$ \\
      \hline
      Class $4$ & $0$ & $0$ & $5$ & $20$ \\
      \hline
\end{tabular}
\caption{Confusion Matrix obtained for Information Gain splitting algorithm}
\label{table-confusion-matrix}
\end{table*}

\section{Decision Tree Visualization and Analysis}
\label{decisionTreeVisualization}
\begin{figure*}[!h]
    \centering
    \includegraphics[width=14.5cm, height=11cm]{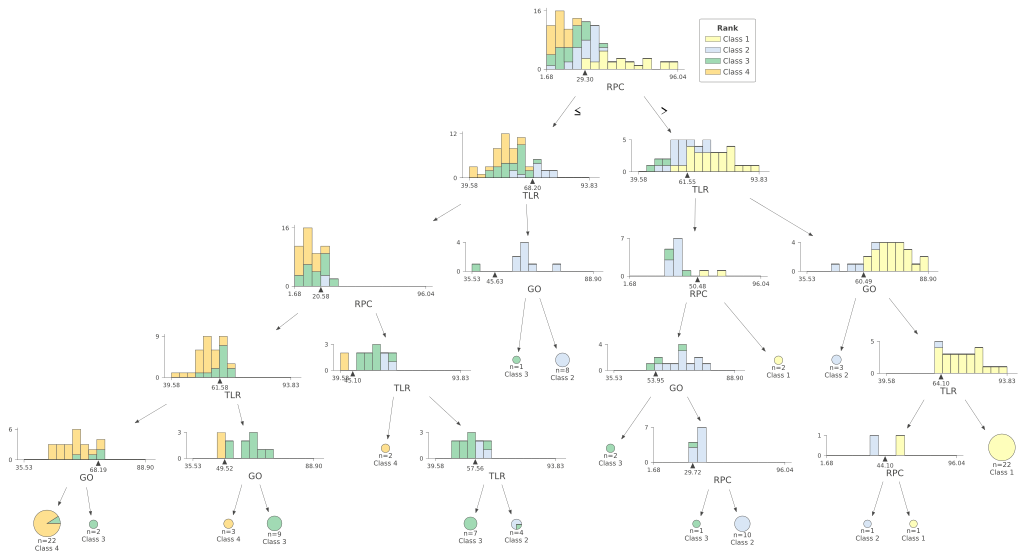}
    \caption{Visualization of Decision Tree for University Ranking Data}
    \label{fig1}
\end{figure*}

The aim of the visualization is to understand how the feature space is split at decision nodes and to look into different paths starting from the root node to leaf nodes (passing through different decision nodes with specific feature and split value). As discussed in Sec. \ref{decision-tree-generation}, the IG splitting algorithm constructs a more promising tree, therefore we visualize the decision tree constructed with the IG splitting algorithm using the dtreeviz python library\cite{fiftynine}. Fig. \ref{fig1} provides a visualization of the DT obtained using NIRF 2018 university ranking data. The splitting of the feature space at each level is shown using black wedges in Fig.~\ref{fig1} and the X-axis depicts the range of values over which the samples are distributed. Figure~\ref{fig1} also aids to visualize feature-target space distributions using histograms. Different colours in the histogram represent different target classes (yellow for Class 1, sky-blue for Class 2, green for Class 3 and orange for Class 4 ). 
The Y-axis height of the stacked histogram is the total number of samples from all classes. The leaf node size is commensurate with the number of samples in that leaf. Leaves with a huge majority target class are very reliable predictors.
Decision nodes lower in the tree, that is as levels of the tree increases, are more and more pure. A pie chart is used for the classifier leaves, a single strong majority category using a particular color gives the leaf prediction.
The key to arrive at a strategy is to examine the decisions taken along the path from the root node to the leaf predictor node.

%As discussed in Sec. \ref{decision-tree-generation}, the IG splitting algorithm constructs a more promising tree that takes into account trends in data of year $2018$ and $2019$. Thus, we visualize the decision tree constructed with the IG splitting algorithm using the dtreeviz python library~\cite{fiftynine}. Fig. \ref{fig1} provides a visualization of the tree. 
In any decision tree, as the root node has the highest entropy value at the point of split, we can state that the root node best classifies the data. As observed in our case, in fig.~\ref{fig1}, RPC is the splitting parameter at the first decision node of the tree (level 1) i.e, it is the parameter that best splits the data and hence it plays the most important role in determining the ranking class of the university. The splitting value of RPC in this case (first decision node) is $29.3$. The majority of samples in the left part of the tree (having RPC values less than $29.3$) belong to Class $4$ and Class $3$, and the majority of the samples in the right part of the tree (having RPC values greater than $29.3$) belong to Class $2$ and Class $1$. Thus, institutes with RPC less than $29.30$ are more likely to fall in class $3$ and $4$ i.e, rank in the $51-100$ range. On the other hand, RPC value greater than $29.30$ suggests that the university is more likely to fall in class $1$ and $2$ i.e, rank in the $1-50$ range. As $29.30$ is very close to the median of RPC, shown using boxplot in Fig. \ref{fig6}, we can infer that institutes having their RPC value in the 3rd and 4th quartile are more likely to be in the top 50 universities i.e, class $1$ and $2$. 
We can observe that at level 1, samples are divided into two parts, left subset  and right subset, based on splitting criteria. The same process is repeated for these subsets at the next level, i.e. take the left path if specific feature in test vector is less than the split point, or else take the right path. In the second level, TLR is the splitting parameter for both the subsets. As we move further down, i.e. decision nodes lower in the tree, we can box in narrower regions of the feature space that are more and more pure. Another interesting observation in this case is that, although the input features in the training set as well as testing set include TLR, RPC, GO and OI,  however, in the Decision Tree visualized in Fig.~\ref{fig1}, we observe that OI is not present in any of the decision nodes. Hence, we can infer that OI is not playing a major role in classifying the universities in different classes in $2018$ and a similar trend is followed in $2019$. In summary, this visualization clearly shows how the model training arrives at a specific tree using the feature-space splits of the decision nodes. As discussed in Sec. \ref{sec8}, deciding the combination of parameters to improve upon for overall score improvement is a major challenge, however visualizing the decision tree as explained above can help a university in taking informed decisions based on existing trends in ranking data.

Using this DT based approach, an institute can estimate the probability of being placed in a particular desired class while following different paths in terms of scores achieved in different parameters. One way to interpret the Decision tree is to understand the paths followed to reach a leaf node classifying the data points into one class. For example, consider the rightmost path $RPC > 29.30$ and $TLR > 64.10$ and $GO > 60.49$, which leads to the rightmost leaf node of the decision tree which has $22$ samples of Class $1$. We can analyse this path to understand the trend of the data that is followed in the year $2018$ and speculate that a similar path is to be followed in $2019$. Using the conditions in the rightmost path, we can state that given these conditions ($RPC > 29.30$ and $TLR > 64.10$ and $GO > 60.49$) are followed a university is very likely to be in class $1$. However, as the score is dependent on all the parameters, we also find the minimum value of other parameters that also needs to be maintained, which can be obtained directly using the data. For the path mentioned above, we apply the filter $RPC > 29.30$ and $TLR > 64.10$ and $GO > 60.49$ on the data and obtain the minimum OI present in the filtered data. In this case, for OI the minimum value of $40.27$ needs to be maintained by a university to be in class $1$. It is also necessary to take the final score into account, i.e improve the RPC, TLR, GO and OI with the above conditions, such that $score \geq 49.559$ ($49.559$ is the minimum score obtained by institutes present in class $1$). Another example is the leftmost path of the tree, which is defined using less than equal to ($\leq$) a specific score. For this path, we must define the minimum value to be maintained for every parameter using the data as the path sets no lower limit on the data. For the leftmost path in the tree, we get $TLR \leq 61.58$ from the decision tree and using the data we define a minimum for TLR i.e it should be greater than the minimum TLR of class $4$, $TLR > 39.58$. Based on Fig. \ref{fig6}, TLR values can be present in the second and first quartile of the data. Hence, the leftmost path can be understood by creating ranges for every feature where the minimum to be maintained is obtained using the data and the maximum of the range can be obtained from the decision tree such that the sum of all values of features obtains a score greater than $33.46$, which is the minimum score of class $4$. It is important to note that considering the minimum may not lead to the score value desired i.e in the example provided $score > 33.46$ and the minimum defined is just the bare minimum that is present in the data. We recommend maintaining an average of the minimum and maximum of the range value for every feature to obtain the required score.  
It is important to quantify the certainty associated with a particular path before planning a particular decision. We can calculate this probability of an institute getting into a certain class at every level and path of the decision tree based on $2018$ year data. Considering, the conditions remain similar next year, these probabilities can help determine a path that can be taken by an institute to improve its ranking. A table showing the probability of an institute being in class 1 at different levels in the rightmost path are presented in Table \ref{tab40}. As extreme probabilities of 0 and 1 that are obtained using maximum likelihood estimate are not reasonable, we add a Laplace correction to the probability estimate~\cite{fifty-seven}. The new probability is determined using Eq. \ref{equation-laplace-estimate}
\begin{equation}
\label{equation-laplace-estimate}
    P(c/x)= \frac{\sum_{i=1}^{n}\delta(c_i,c)+1}{n+n_c}
\end{equation}

$P(c/x)$ determines the probability of an unknown instance being classified as present in class $c$; $n$ is the number of instances in the leaf node(s); $n_c$ is the number of classes; $\delta(c_i,c)$ is one if $c_i=c$ and $0$ otherwise. Considering the rightmost path (last row of Table \ref{tab40}), we can observe that even if the university maintains RPC, TLR, GO a little over the median (as can be seen in Fig. \ref{fig6}) such that the final $score \geq 49.559$ , university has a very good probability of 0.88 to be present in class $1$. Similar table showing the Laplace estimate for the leftmost path for being in class $4$ can be seen in Table \ref{table_class4}.
\begin{table*}[!h]
    \centering
    \begin{tabular}{|c|c|c|c|c|c|c|}
    \hline
    Level & Class & Path & Laplace estimate \\
    \hline
     $1$ & $1$ & \vtop{\hbox{\strut $RPC > 29.3$}\hbox {\strut ($TLR \geq 43.7$ \& $GO \geq 47.12$}\hbox{\strut \& $OI \geq 29.09$)}} & $0.57(26/46)$ \\
     
    \hline
    $2$ & $1$ & \vtop{\hbox{\strut $RPC > 29.3$ \& $TLR > 61.55$}\hbox {\strut ($GO \geq 47.12$ \& $OI \geq 36.54$)}} & $0.77(24/31)$\\
    
    \hline
     $3$ & $1$ & \vtop{\hbox{\strut $RPC > 29.3$ \& $TLR > 61.55$}\hbox{\strut \& $GO > 60.49$}\hbox {\strut ($OI \geq 40.27$)}} & $0.86(24/28)$\\
     
    \hline
    $4$ & $1$ & \vtop{\hbox{\strut $RPC>29.3$ \& $TLR > 64.10$}\hbox{\strut \& $GO > 60.49$}\hbox {\strut ($OI \geq 40.27$))}} & $0.88(23/26)$ \\
    
    \hline

    \end{tabular}
    \caption{Level wise probability (with Laplace correction) of being in class 1.  \\ 
    Score $> 49.559$; Median Values : TLR - $61.295$; RPC - $26.445$; GO - $62.22$; OI - $51.995$ \\ }
    \label{tab40}
\end{table*}

We can observe from Fig~\ref{fig1}, there are three paths that may lead to class 1 (complete yellow leaf nodes with 1,2 and 22 samples totalling 25 samples in class 1). In addition to the rightmost path, another path in the decision tree ($RPC > 50.48$ and $TLR < 61.55$ but TLR greater than a minimum value i.e $TLR \geq 54.64 $) leading to class 1 seen in Fig. \ref{fig6}, has RPC value way above the median although the TLR to be maintained is near the median. In the short run, this may be a daunting task for an institute which is not very strong in terms of research and innovation. Hence, based on the probability of the path, strength and weakness, scopes of improvement and resources at hand, universities can determine the most suitable paths.

\begin{figure*}[!h]
    \centering
    \includegraphics[width=11cm, height=12cm]{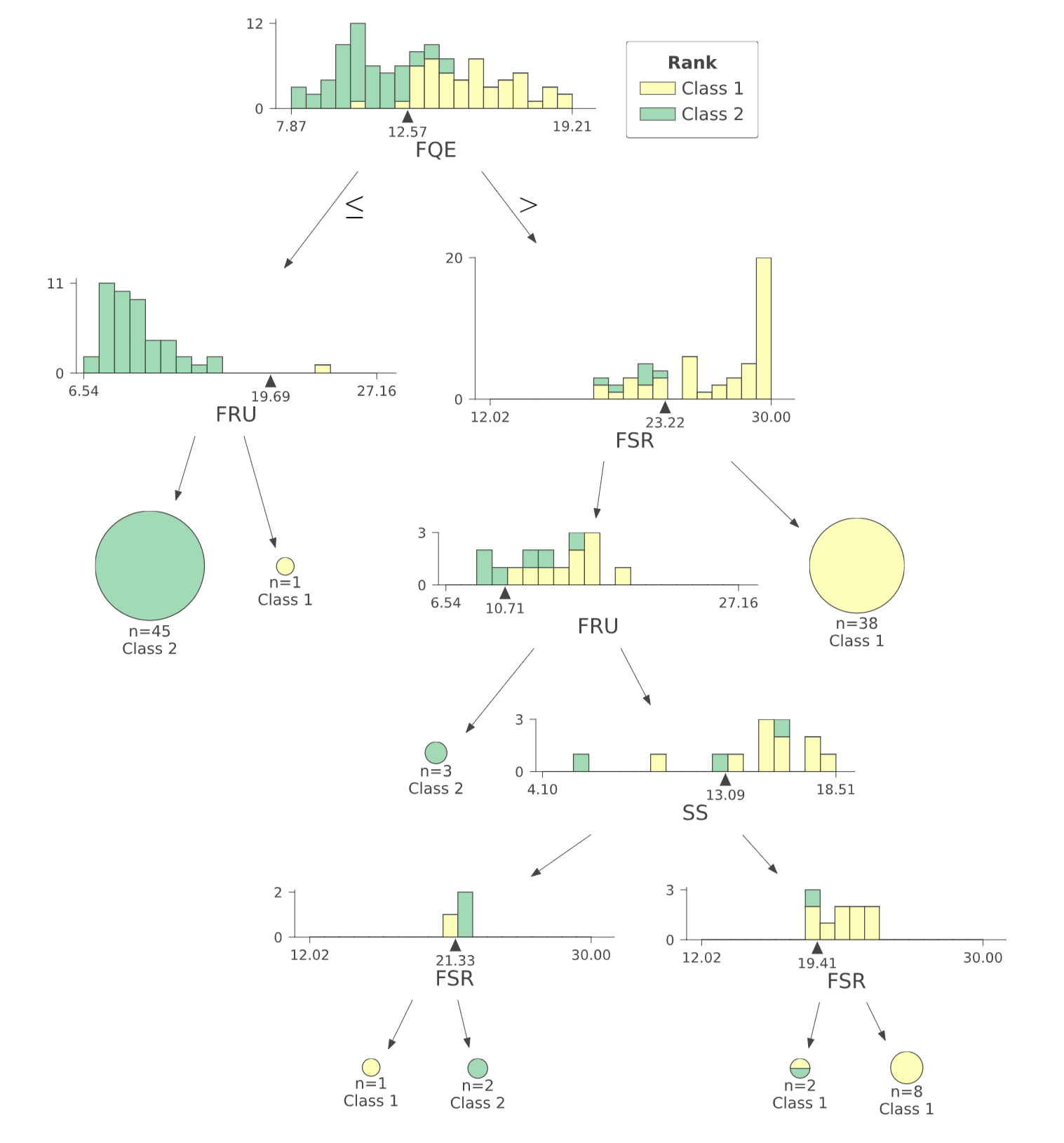}
    \caption{TLR Level - 2 tree visualization \\
    FRU-final resources \& utility, FQE-faculty metric, \\
    SS-student strength, FSR-Faculty Student Ratio,}
    \label{fig2}
\end{figure*}

Further analysis can be performed by constructing a decision tree with similar experimental conditions on sub-parameters used to calculate TLR, RPC, GO and OI. At first, with the help of the decision tree visualized in Fig. \ref{fig1} and box-plots presented in section \ref{section-box-plot}, an institute can find out which first-level parameters to focus on for improvement. To reach a certain value of primary parameter (TLR or RPC or GO or OI), strategies can be prepared based on decision trees created using sub-parameters used to calculate the values of primary parameters. We construct the decision tree for the value of $TLR \geq 61.44$ and visualize it in Fig. \ref{fig2}. This decision tree provides further fine-grained insights about the sub-parameters to focus upon to improve TLR score. We observe that to increase TLR above $61.44$; an institute should work towards increasing the FQE value (quality  of Faculty with experience) as visualized in the first decision node in Fig. \ref{fig2}, and it should also pay attention to FSR (Faculty  Student Ratio). A similar decision tree can be constructed for every decision node presented in Fig. \ref{fig1} to prepare strategies to move from the left path of the decision node to the right path leading to rank improvements in subsequent years.

\begin{table*}[!h]
    \centering
    \begin{tabular}{|c|c|c|c|c|c|c|}
    \hline
    Level & Class & Path & Laplace Estimate \\
    %\hline
     %$1$ & $4$ & $RPC \leq 29.3$ ($TLR \geq 39.58$ \& $GO \geq 35.43$ \& $OI \geq 34.12$)  & $0.42 (26/62)$\\
     
    \hline
    $2$ & $4$ & \vtop{\hbox{\strut $1.68 \leq RPC \leq 29.3$}\hbox{\strut \& $39.58 \leq TLR \leq 68.20$}\hbox{\strut ( $GO \geq 38.77$ \& $OI \geq 34.12$)}} & $0.49 (26/53)$\\
    
    \hline
     $3$ & $4$ & \vtop{\hbox{\strut $1.68 \leq RPC \leq 20.58$}\hbox{\strut \& $39.58 \leq TLR \leq 68.20$}\hbox{\strut ($GO \geq 46.73$ \& $OI \geq 34.12$)}} & $0.6 (24/40)$\\
     
    \hline
    $4$ & $4$ & \vtop{\hbox{\strut $1.68 \leq RPC \leq 20.58$}\hbox{\strut \& $39.58 \leq TLR \leq 61.58$}\hbox{\strut ($GO \geq 46.76$ \& $OI \geq 37.26$)}} & $0.75 (21/28)$\\
    
    \hline
    $5$ & $4$ & \multicolumn{1}{l|}{\vtop{\hbox{\strut $1.68 \leq RPC \leq 20.58$}\hbox{\strut \& $39.58 \leq TLR \leq 61.58$}\hbox{\strut \& $35.43 \leq GO \leq 68.19$}\hbox{\strut  ($OI \geq 37.26$)}}} & $0.81 (21/26)$\\
    
    \hline
    
    \end{tabular}
    \caption{Level-wise Probability (with Laplace correction) of being  in class 4.  \\ 
    Score $>$ 33.46; Median Values : TLR - $61.295$; RPC - $26.445$; GO - $62.22$; OI - $51.995$ }
%    Minimum Values : TLR - $39.58$ ; RPC - $1.68$; GO - $35.43$ ; OI - $29.09$ \\
%    Median: TLR - $54.9$ ; RPC - $11.68$; GO - $56.8$}
    \label{table_class4}
\end{table*}

\section{Conclusions}
University rankings provide a good source of comparative information to various stakeholders and with the increasing importance of rankings year by year, universities have a strong motivation to be ranked amongst the best. However, preparing a quantitative roadmap for rank improvement is a challenging task because the dynamics of higher educational institute rankings is a complex topic involving several variables and parameters such as teaching, research performance, graduation outcome, reputation, resources, outlook,  etc.
A clear understanding of a ranking model which condenses these multivariate data into a single score and analysis of these complex sets of multidimensional information is a prerequisite for rank improvement. \\
In this work, firstly, using NIRF ranking data, we discuss the results of exploratory data analysis methods such as correlation heatmap analysis and box plots. Secondly, we propose a novel way of analyzing university ranking data using decision trees, an interpretable machine learning model. Finally, we show that an appropriate visualization of the decision tree outcomes coupled with probability estimates can provide valuable insights while taking decisions for rank improvements.
 Correlation heatmap analysis on the NIRF ranking data suggests that there is a weak correlation between the parameters considered for ranking, and improving one parameter does not guarantee an improvement in another parameter. Analysis using box plots helps to acquire the first empirical understanding of scope of improvement in different parameters and where does an institute stand with respect to other participating institutes in different parameters. However, for overall rank improvement, an institute needs to meticulously plan the combination of various parameters to improve upon based on availability of resources. To address this issue, we propose the use of decision trees to capture the general trends in the data followed by universities with certain range of ranks. Using predictive analysis, we conclude that the general trends of ranking data remain the same over the years. We provide a detailed analysis of a decision tree constructed using NIRF data and show how different decision paths with different uncertainties may lead to a particular class or quartiles based on range of ranks. Many universities aim to be a part of top $50$ or top $100$ universities in future rankings, the proposed data visualization based strategization for rank improvement using decision tress provides a novel and quantitative approach towards that journey.
Our study also shows that it is critical for the ranking organizations to make the complete ranking data public which will help universities to prepare data-driven strategies for improvement in different parameters and will also help the public to obtain the most objective picture of the position of particular HEIs in relation to one another.

% \bibliography{sn-bibliography}
%\begin{verbatim}
\bibliographystyle{abbrv}
\bibliography{article}
\end{document}